\documentclass[11pt]{article}
\usepackage{coling2020}
\usepackage{times}
\usepackage{latexsym}
\usepackage{amssymb}
\usepackage{url}
\usepackage{examples}
\usepackage{multirow}
\usepackage{color}
\usepackage{xcolor,colortbl}
\usepackage{graphicx}
\usepackage{amsmath}
\definecolor{lightgrey}{RGB}{227,226,219}

\usepackage{microtype}

\colingfinalcopy 

\title{
Fine-grained Information Status Classification Using Discourse Context-Aware BERT
}

 \author{
Yufang Hou \\
 IBM Research Europe, Ireland\\ {\tt yhou@ie.ibm.com}\\
    }

\date{}

\begin{document}
\maketitle

\begin{abstract}
Previous work on bridging anaphora recognition \cite{houyufang13b} 
casts the problem as a subtask of learning fine-grained information status (IS). However, 
these systems heavily depend on many hand-crafted linguistic features. In this paper,
we propose a simple discourse context-aware BERT model for fine-grained IS classification. On the ISNotes corpus \cite{markert12}, our model 
achieves new state-of-the-art performance on fine-grained IS classification, obtaining a 4.8 absolute overall accuracy improvement compared to \newcite{houyufang13b}. More importantly, we also show an improvement 
of 10.5 F1 points for 
bridging anaphora recognition without using any complex hand-crafted semantic features designed for capturing the bridging phenomenon.
We further analyze the trained model and find that the most attended signals for each IS category correspond well to linguistic notions of information status.
\end{abstract}

 \section{Introduction}
\label{sec:intro}
\emph{Information Structure} \cite{halliday67,prince81a,prince92,gundel93,lambrecht94,birner98,kruijff-korbayova03}
studies structural and semantic properties of a sentence according to its relation to the 
discourse context. Information structure affects how discourse entities are referred to in a
 text, which
is known as \emph{Information Status} \cite{halliday67,prince81a,nissim04}.
Specifically, information status (IS henceforth) reflects the accessibility of a discourse entity
based on the evolving discourse context and the speaker's assumption about the hearer's knowledge and beliefs.
For instance, according to \newcite{markert12}, \emph{old} mentions\footnote{A mention is a noun phrase which refers to a discourse entity and carries information status.} refer to entities that 
have been referred to previously; \emph{mediated} mentions have not
been mentioned before but are accessible to the hearer by reference to another \emph{old} mention or to prior world knowledge; and \emph{new} mentions refer to entities that are introduced to the discourse for the first time and are not known to the hearer before.

In this paper, we mainly follow the IS scheme proposed by \newcite{markert12} and focus on learning fine-grained IS on written texts.  
A mention's semantic and syntactic properties can signal its information status. For instance, indefinite NPs tend to be \emph{new} and pronouns are likely to be \emph{old}. 
 Moreover, referential patterns of how a mention is referred to in a sentence also affect this mention's IS. 
In Example \ref{exa:exa11},  
``Friends'' is a bridging anaphor even if we do not know the antecedent (i.e., \emph{she}); while the information status for ``Friends'' in Example \ref{exa:exa12} is \emph{mediated/worldKnowledge}. Section \ref{subsec:motivation} analyzes the characteristics of each IS category and 
the relations between IS and discourse context.

\begin{examples}
\item \label{exa:exa11} [\emph{She}]$_{antecedent}$ made money, but spent more. {\bfseries Friends} pitched in.
\item \label{exa:exa12} \underline{Friends} are part of the glue that holds life and faith together.
\end{examples}

In this work, we propose a simple yet effective discourse context-aware self-attention model based on BERT \cite{devlin2018bert} for fine-grained IS classification. We find that the sentence containing the target mention as well as the lexical overlap information between the target mention and the preceding mentions are the most important discourse context when assigning IS for a mention. With the self-attention mechanism,  our model can capture important signals within a mention and the interactions between the mention and its context. On the ISNotes corpus \cite{markert12}, our model achieves new state-of-the-art performance on fine-grained IS classification, obtaining a 4.8 absolute overall accuracy improvement compared to \newcite{houyufang13b}. More importantly, we also show an improvement 
of 10.5 F1 points for 
bridging anaphora recognition without using any sophisticated hand-crafted semantic features.

Furthermore, to gain additional insights into our model's predictions, we analyze the attention mechanisms of our trained model. We find that the most attended tokens for each IS category correspond well with linguistic features of information status. For instance, for the \emph{old} IS category, the most attended token list includes pronouns such as ``she'', ``her'',  and ``it''. While for the \emph{new} category, the model pays more attention to indefinite determiners such as ``a'' and ``an''. Section \ref{sec:attention} provides a detailed analysis of the attention map for each IS category.

To summarize, the main contributions of our work are as follows: 
\begin{itemize}
\item We propose a simple and effective model for fine-grained IS classification. The model uses a novel approach for encoding information from the previous sentences along with the current sentence for IS classification.
\item  Our proposed model achieves new state-of-the-art results for IS classification and bridging anaphora recognition on the ISNotes corpus. Our model also achieves competitive results for fine-grained IS classification on the Switchboard dialogue IS corpus \cite{nissim04} that uses a different IS scheme than the one in ISNotes. The  processed  datasets  and  code  are
publicly available at: \url{https://github.com/IBM/bridging-resolution}.
\item We carry out ablation studies to understand the effectiveness of each component in our model. We further investigate the self-attention patterns in our model and find that the model does learn specific linguistic features for predicting information status.
\end{itemize}

\section{Related Work}
\label{sec:relatedwork}
\paragraph{IS classification and bridging anaphora recognition.}
Bridging resolution \cite{houyufang14,houyufang18c} contains two sub tasks: identifying bridging anaphors \cite{markert12,houyufang13b,houyufang16} and finding the correct antecedents among candidates \cite{houyufang13a,houyufang18b,houyufang18,houyufang20a}. Most previous studies handle bridging anaphora recognition as part of IS classification problem.
\newcite{markert12} applied joint inference for IS classification on the ISNotes corpus but reported very low results on bridging recognition.
Building on this work, \newcite{houyufang13b} designed many linguistic features to capture bridging and integrated them into a cascading collective classification algorithm. 
This approach later was integrated into a pipeline for bridging resolution \cite{houyufang16b,houyufang18c}.
Differently, \newcite{houyufang16} used an attention-based LSTM model based on GloVe vectors and a small set of features for IS classification. The author reported similar results as \newcite{houyufang13b} regarding the overall IS classification accuracy but the result on bridging anaphora recognition is much worse than \newcite{houyufang13b}.

\newcite{rahman12} incorporated carefully designed rules into an 
SVM algorithm for IS classification on the Switchboard dialogue IS corpus \cite{nissim04}.\footnote{It is worth noting that bridging antecedent information was not annotated in Switchboard. Also, bridging anaphora annotation in Switchboard includes non-anaphoric cases.}  
The authors first designed a rule-based system to assign IS classes to mentions on the basis of Nissim’s IS annotation guidelines \cite{nissim04}. They then applied an SVM$^{multiclass}$ algorithm for this task by combining the prediction from the rule-based system, the ordering of the rules as well as two lexical features.

Another work on IS classification was carried out by \newcite{cahill12}.  They assumed that the distribution of IS classes within sentences tends to have certain linear patterns, e.g., \emph{old} $>$ \emph{mediated} $>$ \emph{new}. Under this assumption, they trained a CRF model with syntactic and surface features for fine-grained IS classification on the German DIRNDL radio news corpus \cite{riester10}. Recently, \newcite{roesiger19} adapted eight rules from \newcite{houyufang14} to recognize bridging anaphors and find their antecedents in the improved annotations of the extended DIRNDL corpus \cite{bjoerkelund14b}.

Different from the above-mentioned work, we do not use any complicated hand-crafted features, and our model improves the previous state-of-the-art results 
on both overall IS classification accuracy and bridging recognition 
by a large margin on the ISNotes corpus. Our model also achieves competitive results 
for fine-grained IS classification on Switchboard compared to the approach in \newcite{rahman12} that uses the Stanford coreference resolver and an SVM classifier that explores 18 carefully designed hand-crafted rules.

\paragraph{Fine-tuning with contextual word embeddings.}
Recent studies 
\cite{peter18,devlin2018bert}
 have shown that a range of downstream NLP tasks benefit from fine-tuning task-specific parameters with pre-trained contextual word representations. Our work belongs to this category and we fine-tune our model based on 
BERT representations \cite{devlin2018bert}.
The novelty of our approach is that we create a ``pseudo sentence'' for each mention that encodes the most effective local and global discourse context for predicting the mention's IS. The self-attention mechanism in Transformer's self-attention encoder \cite{ashish17} allows our model to attend to both the context and the mention itself for clues that are helpful 
to predict the mention's IS.

\paragraph{Model probing.}
Recently,  there have been a number of studies exploring the types of knowledge encoded in the BERT model. \newcite{jawahar-etal-2019-bert} found that internal vector representations in BERT encode rich linguistic information, with surface information at the bottom layers, syntactic information in the middle layers, and semantic information at the top layers. \newcite{clark-etal-2019-bert} showed that certain attention heads in BERT correspond well to the linguistic knowledge of syntax and coreference. In our work, we demonstrate that 
the attention patterns in our trained model embed linguistic notions of information status.

\section{Approach}
\label{sec:method}

\subsection{Information Status and Discourse Context}
\label{subsec:motivation}
The IS scheme proposed by \newcite{markert12} adopts three major course-grained IS categories (\emph{old}, \emph{new}, 
and \emph{mediated}) from \newcite{nissim04} and distinguishes six subcategories for \emph{mediated}. 
Below we provide a brief description for the eight fine-grained IS classes in ISNotes.

\emph{Old} mentions are coreferent with the already introduced entities. \emph{New} 
mentions are entities that have not been introduced into the discourse and the hearer/reader cannot infer them from either previously mentioned entities or general world knowledge. 
\emph{Mediated} mentions are discourse-new and hearer-old \cite{prince92}. They have not been introduced into the discourse before but are accessible to the hearer by reference to another mention or to prior world knowledge. 

Among the \emph{mediated} category, \emph{Mediated/worldKnowledge} mentions are generally known to the hearer. This category contains mostly proper names. \emph{Mediated/syntactic} mentions are syntactically linked to other \emph{old} or \emph{mediated} mentions, such as ``[[their]$_{old}$ father]$_{m/syntactic}$'' or ``[a war in [Africa]$_{mediated}$]$_{m/syntactic}$''. \emph{Mediated/aggregate} mentions are coordinated NPs where at least one element is \emph{old} or \emph{mediated}, such as ``[[U.S.]$_{mediated}$ and [Canada]$_{mediated}$]$_{m/aggregate}$''. \emph{Mediated/function} mentions refer to a value of a previously explicitly mentioned function and this function needs to be able to rise or fall (e.g., \textbf{6 cents} in Example \ref{exa:exa_function}). \emph{Mediated/comparative} mentions usually contain a premodifier to indicate that this entity is compared to another preceding entity (antecedent) (e.g., \textbf{further attacks} in Example \ref{exa:exa_comparative}). Finally,  \emph{Mediated/bridging} mentions are associative anaphors that link to previously introduced related entities/events (e.g., \textbf{Friends} in Example \ref{exa:exa11}).

\begin{table*}[t]
\begin{footnotesize}
\begin{tabular} {|l|l|l|c|c|c|}
\hline
&&& \multicolumn{3}{c|}{Factors affecting IS} \\
 &Description  & Example & Mention  & Local  & Previous \\
 &  &  & Properties &Context & Context\\
  \hline
 {old} &coreferent with an already introduced entity&\emph{he}, \emph{the president}& \checkmark &\checkmark &\checkmark \\ \hline
 {m/worldKnow.} &generally known to the hearer&\emph{Francis}, \emph{the pope}&\checkmark &\checkmark &\\ \hline
 {m/syntactic} &syntactically linked to other  \emph{old} or \emph{mediated} &\emph{their father}&\checkmark &&\\ 
 &mentions&\emph{a war in Africa}& &&\\ \hline
 {m/aggregate} &coordinated NPs where at least one element 
  &\emph{U.S. and Canada}&\checkmark & &\\ 
 &is \emph{old} or \emph{mediated} &\emph{he and his son}& & &\\ \hline
 {m/function} &refer to a value of a previously explicitly  &(the price went&\checkmark &\checkmark &\\ 
 &mentioned rise/fall function&down) \emph{6 cents}& & &\\ \hline
 {m/comparative} &usually contain a premodifier to indicate that &\emph{another law}&\checkmark &\checkmark & \checkmark \\ 
  & this entity is compared to another entity &\emph{further attacks}& & &  \\ \hline
 {m/bridging} & associative anaphors which link to previously &\emph{the price}&\checkmark &\checkmark & \checkmark \\ 
 & introduced related entities/events&\emph{the reason}& & & \\ \hline
 {new} &introduced into the discourse for the first time &\emph{a reader}&\checkmark &\checkmark & \checkmark \\  
&and not known to the hearer before&\emph{politics}&&& \\  \hline
\end{tabular}
\end{footnotesize}
\caption{Information status categories and their main affecting factors. ``Local context'' means 
the sentence $s$ which contains the target mention, ``Previous context'' indicates all sentences from the discourse which occur before $s$.}
\label{tab:motivation}
\end{table*}

\begin{examples}

\item \label{exa:exa_function} In trading on the American Stock Exchange, Delmed's price [went down]$_{function}$ \textbf{6 cents}. 
\item \label{exa:exa_comparative} [\emph{The cyber attacks}]$_{antecedent}$ were followed by \textbf{further attacks} on ZDNet.com, a news portal. 
\end{examples} 

We characterize the linguistic factors that affect a mention's IS into three categories: mention properties, local context, as well as previous context. 
Table \ref{tab:motivation} lists the definitions for these IS categories
and summarizes the main affecting factors for each IS class. Note that we analyze the main affecting factors for each IS class based on their definitions.

As described in Section \ref{sec:intro}, a mention's internal syntactic and semantic properties can signal its IS. For instance, a mention containing
a possessive pronoun modifier is likely to be \emph{mediated/syntactic} (e.g., \emph{their father}); and a \emph{mediated/comparative} mention often contains a premodifier indicating that this entity is compared to another preceding entity (e.g., \emph{further attacks}).

In addition, for some IS classes, the ``local context'' (the sentence $s$ which contains the target mention) and 
``previous context'' (sentences from the discourse which precede $s$) play an important role when assigning IS to a mention. Example \ref{exa:exa11} and Example \ref{exa:exa12} in Section \ref{sec:intro} demonstrate the role of the local context for IS. 
In Example \ref{exa:exa11}, the referential patterns in the local context indicate that ``Friends'' is a bridging anaphor,\footnote{\newcite{clarkherberth75} uses necessary, probable and inducible parts/roles to distinguish different types of bridging and argues that only in the first case the antecedent triggers the bridging anaphor in the sense that we already spontaneously think of the anaphor when we read/hear the antecedent. In the probable/inducible cases, the bridging anaphora accommodates itself into the context and is induced by the need for an antecedent. Section \ref{sec:ablation} illustrates this using a wug-test example.} whereas ``Friends'' in Example \ref{exa:exa12} is a generic NP.

 Sometimes we need to look at the previous context when deciding IS for a mention. In Example \ref{exa:exa4}, without looking at the previous context, we tend to think the IS for ``Poland'' in the second sentence is \emph{mediated/WorldKnowledge}. Here the correct IS for  ``Poland'' is \emph{old} because it is mentioned before in the previous context.

\begin{examples}
\item \label{exa:exa4} [Previous context:]  In \textbf{Poland}, only 4\% of all investment goes toward making things farmers want; in the West, it is closer to 20\%. \newline [Local context:]  A private farmer in \textbf{Poland} is free to buy and sell land.
\end{examples}

\subsection{IS Classification with Discourse Context-Aware Self-Attention}
\label{subsec:model}
To account for the different factors described in the previous section when predicting IS for a mention, we create a novel ``pseudo sentence'' for each mention and apply the multi-head self-attention encoder \cite{ashish17,devlin2018bert} to this sentence. 

\begin{figure*}[t]
\begin{center}
\includegraphics[width=1.0\textwidth]{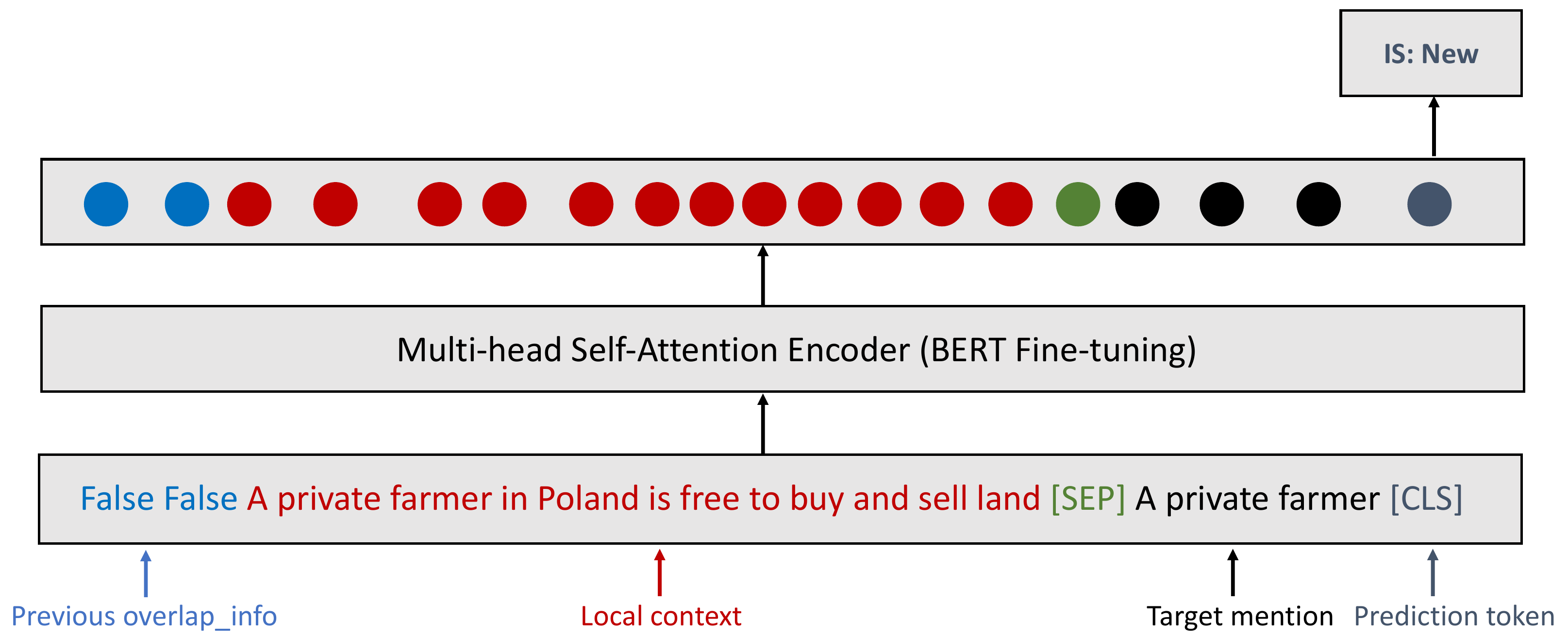}
\end{center}
\caption{Fine-grained IS classification with discourse context-aware self-attention.}
\label{fig:model}
\end{figure*}

Figure \ref{fig:model} depicts the high-level structure of our model. The pseudo sentence consists of five parts: previous overlap\_info, local context, the delimiter token ``[SEP]'', the content of the target mention, and the IS prediction token ``[CLS]''. The previous overlap\_info part contains two tokens, which indicate 
whether the target mention has the same string/head with a mention from the preceding sentences. And the local context is the sentence containing the target mention. 

The final prediction is made based on the hidden state of the prediction token ``[CLS]''. 
In principle, this is similar to BERT's ``[CLS]a[SEP]b'' framework, in which the special classification token (\emph{[CLS]}) is added to every sequence as the first token and its hidden state is used as the aggregate sequence representation for classification tasks. The novelty of our work is that we design the structure of \emph{a} and \emph{b} in a way that embeds the most indicative information to predict a mention's information status. During the training stage, the mechanism of multi-head self-attention helps the model to learn the important cues from both the mention itself and its discourse context when predicting IS.

There are other ways to encode a mention's context information. For instance, one could try to add more previous sentences in the local context or replace the current \emph{previous overlap\_info} with all previous sentences. In practice, we found that the current configuration 
yields the best results on the ISNotes and Switchboard IS corpora. In particular, we notice that
using all previous sentences  as the discourse context significantly decreases the results for IS classification.\footnote{When the pseudo sentence is longer than 512 tokens, we use a sliding window of 100 tokens to account for all previous context of a mention.} 
This is in line with the observation from \newcite{joshi-etal-2019-bert} that modeling the longer context in BERT provides no improvement for coreference resolution.

\subsection{Model Parameters}
\label{subsec:parameter}
We use the vanilla BERT for our experiments.
We initialize our model using 
pre-trained BERT contextual embeddings, which is trained on top of the BookCorpus (800M words) and English Wikipedia (2,500M words). 
 We then fine-tune the model for 3 epochs with the learning rate of $3e-5$ and a batch size of 32. During training and testing,  the max token size of the pseudo sentence is set as 128.\footnote{The length of 128 tokens covers all the cases in ISNotes. In practice, we truncate the local sentence if the whole pseudo sentence is longer than 128 tokens.}

\section{Experiments on ISNotes}

\subsection{Experimental Setup}
We perform experiments on the ISNotes corpus \cite{markert12}, which contains 10,980 mentions annotated for information status
in 50 news texts
taken from the Wall Street Journal portion of the OntoNotes corpus \cite{ontonotes4.0data}. 
Table \ref{tab:is} shows the IS distribution in ISNotes.

\begin{table}[h]
\begin{center}
\begin{tabular}{lrrr}
Mentions & &10,980& \\ \hline\hline
old & & 3,237&29.5\% \\ \hline
mediated & & 3,708&33.8\%\\ \hline
& syntactic & 1,592&14.5\%\\
& world knowledge & 924&8.4\%\\
& bridging & 663&6.0\%\\
& comparative & 253&2.3\%\\ 
& aggregate & 211&1.9\%\\
& func & 65&0.6\%\\
 \hline
new & & 4,035&36.7\%\\
\hline
\end{tabular}
\end{center}
\caption{\label{tab:is}  IS distribution in ISNotes.}
\end{table}

\begin{table*}[t]
\begin{footnotesize}
\begin{center}
\begin{tabular}{|p{1.8cm}|p{0.466cm}p{0.466cm}p{0.466cm}|p{0.466cm}p{0.466cm}p{0.466cm}|p{0.466cm}p{0.466cm}p{0.466cm}|p{0.466cm}p{0.466cm}p{0.466cm}|p{0.466cm}p{0.466cm}p{0.466cm}|}
\hline
 &  \multicolumn{9}{c|}{\emph{baselines}} &\multicolumn{6}{c|}{\emph{this work}} \\ 
 &  \multicolumn{3}{c|}{\emph{collective}} &\multicolumn{3}{c|}{\emph{cascade collective}} & \multicolumn{3}{c|}{\emph{incremental LSTM }} & \multicolumn{3}{c|}{\emph{self-attention with}}& \multicolumn{3}{c|}{\emph{self-attention with}}\\ 
 &  \multicolumn{3}{c|}{Hou et al.(2013)} &\multicolumn{3}{c|}{Hou et al.(2013)} & \multicolumn{3}{c|}{Hou (2016)} & \multicolumn{3}{c|}{\emph{BERT$_{BASE}$}}& \multicolumn{3}{c|}{\emph{BERT$_{LARGE}$}}\\ 
 & R & P & F & R & P & F& R & P & F& R & P & F& R & P & F\\ \hline
\hline
 &&&&&&&&&&&&&&& \\
 {old}&84.4&86.0&85.2&82.2&87.2&84.7&85.4&84.9&85.2&87.8&90.5&89.1&88.4&90.0&\textbf{89.2}\\
 {m/worldKnow.}&67.4&77.3&72.0&67.2&77.2&71.9&67.1&74.5&70.6&74.9&77.8&76.3&77.7&79.5&\textbf{78.6}\\
 {m/syntactic} & 82.2&81.9&82.0&81.6&82.5&82.0&80.8&81.9&81.4&82.9&79.5&81.1&83.7&81.1&\textbf{82.4}\\
 {m/aggregate}&64.5&79.5&71.2&63.5&77.9&70.0&67.8&84.6&75.3&76.8&74.7&75.7&80.1&79.3&\textbf{79.7} \\
 {m/function} & 67.7&72.1&69.8&67.7&72.1&69.8&64.6&76.4&70.0&35.9&62.2&45.5&73.4&85.5&\textbf{79.0}\\
 {m/comparative} & 81.8&82.1&82.0&86.6&78.2&82.2&77.9&83.1&80.4&88.1&84.8&86.4&90.5&86.7&\textbf{88.6}\\
 {m/bridging}& 19.3&39.0&25.8&44.9&39.8&42.2&15.7&32.3&21.1&43.3&49.6&46.2&51.0&54.5&\textbf{52.7}\\
 {new} & 86.5&76.1&81.0&83.0&78.1&80.5&87.2&74.8&80.5&85.7&82.5&84.1&86.6&85.2&\textbf{85.9}\\
 \hline
 acc &\multicolumn{3}{c|}{78.9} & \multicolumn{3}{c|}{78.6} & \multicolumn{3}{c|}{78.6} & \multicolumn{3}{c|}{82.0}&\multicolumn{3}{c|}{\textbf{83.7}}\\
\hline
\end{tabular}
\end{center}
\end{footnotesize}
\caption{Results of the discourse context-aware self-attention model compared to the baselines on ISNotes. Bolded scores indicate the best performance for each IS class. The improvements of \emph{self-attention with BERT$_{BASE}$} and \emph{self-attention with BERT$_{LARGE}$} over the baselines are statistically significant at p$<$0.01 using randomization test.}
\label{tab:results}
\end{table*}

Following \newcite{houyufang13b}, all experiments are performed via 10-fold cross-validation on documents. On each testing fold, the model is trained on the other nine folds. The hyper-parameters of 3 epochs and the learning rate $3e-5$ 
were fixed during all training processes. We report overall accuracy as well as precision, recall and F-score per IS class. In the following, we describe the baselines as well as our model with different settings.

\paragraph{\emph{collective} (baseline 1).} \newcite{houyufang13b} applied collective classification to account for the linguistic relations among IS categories.
 They explored a wide range of features (34 in total), including a large number of lexico-semantic features (for recognizing bridging) as well as a couple of surface features and syntactic features.
 
\paragraph{\emph{cascaded collective} (baseline 2).} This is the cascading minority preference system for bridging anaphora recognition from \newcite{houyufang13b}. 

\paragraph{\emph{incremental LSTM} (baseline 3).} This is the attention-based LSTM model proposed by \newcite{houyufang16}. The model uses one-hot vectors to encode IS classes and predicts information status for all mentions of a document from left to right incrementally. 

\paragraph{\emph{self-attention with BERT$_{BASE}$}.} We fine-tune BERT$_{BASE}$ 
on the pseudo sentences described in Section \ref{sec:method}. The model has 12 transformer blocks, 768 hidden units, and 12 self-attention heads. 

\paragraph{\emph{self-attention with BERT$_{LARGE}$}.} We fine-tune 
BERT$_{LARGE}$ on the pseudo sentences described in Section \ref{sec:method}. The model has 24 transformer blocks, 1024 hidden units, and 16 self-attention heads.

\subsection{Results and Discussion}
Table \ref{tab:results} shows the results of our models compared to the baselines. 
Our best model \emph{self-attention with BERT$_{LARGE}$} improves over all baselines 
by a large margin on all IS categories. It achieves an overall accuracy of 83.7\% on fine-grained IS classification, obtaining a 4.8 and 5.1 absolute improvements in accuracy over the two strong  baselines (\emph{collective} and \emph{cascade collective}), respectively. 

It is worth noting that recognizing bridging anaphora is a challenging task \cite{markert12}. 
\newcite{houyufang13b} proposed a lot of discourse structure, lexico-semantic and genericity detection features to capture the phenomenon. Their best model for bridging anaphora recognition (\emph{cascade collective}) achieves an F-score of 42.2. Overall, our model \emph{self-attention with BERT$_{LARGE}$} achieves the new
state-of-the-art performance for this task with an F-score of 52.7
without resorting to any hand-crafted sophisticated semantic features.  
By comparing the confusion matrices of \emph{cascade collective} and \emph{self-attention with BERT$_{LARGE}$}, we find that the highest proportion of recall errors of bridging recognition in \emph{cascade collective} is due to the fact that a lot of bridging anaphors are misclassified as \emph{new}. This can be explained as the syntactic form of many new mentions and bridging anaphors are the same (see Example \ref{exa:exa11} and Example \ref{exa:exa12} ), the lexico-semantic features in \emph{cascade collective} only pick up on certain types of bridging. In addition, most precision errors in \emph{cascade collective} are \emph{new} and \emph{old} mentions being misclassified as \emph{m/bridging}. Both these recall and precision errors are less frequent in \emph{self-attention with BERT$_{LARGE}$}.
It seems that our model does capture properties of bridging anaphora better by only looking at a mention and its interactions with the surrounding context.

\subsection{Ablations}
\label{sec:ablation}
To better understand the impact of different components in our model, we carry out an ablation experiment. We remove \emph{target mention}, \emph{local context}, \emph{previous overlap\_info}, as well as all context information (\emph{local context} + \emph{previous overlap\_info}) from our best model \emph{self-attention with BERT$_{LARGE}$}, respectively. Table \ref{tab:ablation} reports  the  results  of  different  configurations for our model. 

\begin{table*}[t]
\begin{footnotesize}
\begin{center}
\begin{tabular}{|p{1.8cm}|p{0.466cm}p{0.466cm}p{0.466cm}|p{0.466cm}p{0.466cm}p{0.466cm}|p{0.466cm}p{0.466cm}p{0.466cm}|p{0.466cm}p{0.466cm}p{0.466cm}|p{0.466cm}p{0.466cm}p{0.466cm}|}
\hline
 &  \multicolumn{3}{c|}{\emph{self-attention with}} &\multicolumn{3}{c|}{\emph{self-attention wo}} & \multicolumn{3}{c|}{\emph{self-attention wo}} & \multicolumn{3}{c|}{\emph{self-attention wo}}& \multicolumn{3}{c|}{\emph{self-attention wo}}\\ 
 &  \multicolumn{3}{c|}{\emph{BERT$_{LARGE}$}} &\multicolumn{3}{c|}{\emph{target mention}} & \multicolumn{3}{c|}{\emph{local context}} & \multicolumn{3}{c|}{\emph{pre\_overlap\_info}}& \multicolumn{3}{c|}{\emph{local context}}\\ 
  &  \multicolumn{3}{c|}{} &\multicolumn{3}{c|}{} & \multicolumn{3}{c|}{} & \multicolumn{3}{c|}{}& \multicolumn{3}{c|}{\emph{pre\_overlap\_info}}\\
 & R & P & F & R & P & F& R & P & F& R & P & F&R&P&F\\ \hline
\hline
 &&&&&&&&&&&&&&& \\
 {old} & 88.4&90.0&\textbf{89.2}&78.2&75.8&77.0&87.1&90.8&88.9&80.8&87.4&84.0&74.8&84.9&79.5\\
 {m/worldKnow.} & 77.7&79.5&\textbf{78.6}&15.9&37.7&22.4&75.9&77.6&76.7&70.1&64.1&67.0& 64.7&56.5&60.3\\
 {m/syntactic}&83.7&81.1&82.4&23.8&42.8&30.6&85.1&81.2&83.1&85.6&81.4&\textbf{83.4}&84.6&80.3&82.4\\
 {m/aggregate} & 80.1&79.3&79.7&15.2&38.1&21.7&84.4&78.8&\textbf{81.5}&77.3&81.5&79.3&78.7&78.7&78.7\\
 {m/function}& 73.4&85.5&79.0&43.8&41.8&42.7&62.5&62.5&62.5&75.0&90.6&\textbf{82.1}& 65.6&65.6&65.6\\
 {m/comparative} &90.5&86.7&\textbf{88.6}&6.7&31.5&11.1&90.5&86.7&\textbf{88.6}&89.7&86.0&87.8&88.9&84.3&86.5\\
 {m/bridging} & 51.0&54.5&\textbf{52.7}&4.7&38.3&8.3&43.7&48.9&46.2&51.9&52.5&52.2& 40.3&46.4&43.1\\
{new}&86.6&85.2&\textbf{85.9}&80.9&53.7&64.5&85.1&82.6&83.8&86.5&84.6&85.5&85.1&80.3&82.6\\
 \hline
 acc &\multicolumn{3}{c|}{\textbf{83.7}} & \multicolumn{3}{c|}{58.5} & \multicolumn{3}{c|}{82.3} & \multicolumn{3}{c|}{81.1}& \multicolumn{3}{c|}{77.4}\\
\hline
\end{tabular}
\end{center}
\end{footnotesize}
\caption{Ablation experiments results of \emph{self-attention with BERT$_{LARGE}$}
for fine-grained IS classification. Bolded scores indicate the best performance for each IS class. The differences between \emph{self-attention with BERT$_{LARGE}$} and other variations are statistically significant at p$<$0.01 using randomization test.}
\label{tab:ablation}
\end{table*}

Surprisingly, the model 
considering only the content of mentions 
 (see the last column of Table \ref{tab:ablation}) achieves competitive results as the baseline \emph{cascade collective} which explores many hand-crafted linguistic features. 
Also it outperforms the three baselines on several IS categories (\emph{m/syntactic}, \emph{m/aggregate}, \emph{m/comparative}, \emph{m/bridging} and \emph{new}). In Section \ref{subsec:motivation}, we analyze that \emph{m/syntactic} and 
\emph{m/aggregate} are often signaled by mentions' internal syntactic structures, and that the semantics of certain premodifiers is a strong signal for \emph{m/comparative}. 
The improvements on these categories show that our model can capture the semantic/syntactic properties of a mention when predicting its IS. 

Among all three components, it seems that the content of mentions has the most impact on the overall results, while the local context has the least impact. Furthermore, we find that \emph{local context} and \emph{previous overlap\_info} have different impacts on IS classes. More specifically, we notice that \emph{m/bridging}, \emph{m/function} and \emph{new} benefit most from \emph{local context}, whereas \emph{old} and \emph{m/worldKnowledge} 
benefit most from \emph{previous overlap\_info}. This may seem counter-intuitive for \emph{m/bridging} and \emph{m/worldKnowledge}, as one expects that 
\emph{m/bridging} should benefit more from the previous context and  \emph{m/worldKnowledge} is a local phenomenon. For \emph{m/worldKnowledge}, this is 
explained by the fact that the system without previous context information (\emph{self-attention wo pre\_overlap\_info}) wrongly predicts a lot of
\emph{old} mentions as \emph{m/worldKnowledge}, as illustrated in Example \ref{exa:exa4}. 

For \emph{m/bridging}, the big impact of the local context corresponds to \newcite{houyufang13b}'s observation that some bridging can be indicated by referential patterns without world knowledge about the anaphor/antecedent NPs. For instance, in the following sentence, ``\emph{The blicket couldn’t be connected to the dax. \textbf{The wug} failed.}'', the mention ``\emph{\textbf{The wug}}'' is likely a bridging anaphor, although we do not know the antecedent.\footnote{We thank an anonymous reviewer for bringing up this example in one of our previous work \cite{houyufang13b}. We also thank another anonymous reviewer of this work for pointing out that ``\emph{\textbf{The wug}}'' could also be an epithet.} Similarly, \newcite{clarkherberth75} distinguishes between bridging via necessary, probable, and inducible parts/roles. He states that only in the first case does the antecedent trigger the bridging anaphor in the sense that we already spontaneously think of the anaphor when we read/hear the antecedent.  In the probable/inducible cases,  bridging anaphora accommodates itself into the context and is induced by the need for an antecedent. 

In addition, we also tested whether a broader local context can help us to detect bridging better. 
In the ISNotes corpus, 26\% of bridging anaphors have the antecedents from the same sentence, and 77\% of anaphors have antecedents occurring in the same or
 up to two sentences prior to the anaphor. In practice, we tried to add the previous $k$ sentences  ($k=1$ and  $k=2$) into the current local context 
but found that the overall results for bridging in both settings are similar to the current one. 
 
\section{Experimental Results on the Switchboard IS Corpus}
In this section, we apply our discourse context-aware self-attention model to the Switchboard dialogue IS corpus \cite{nissim04}. The corpus contains around 63k mentions annotated with IS types (i.e., \emph{old}, \emph{mediated}, and \emph{new}) and subtypes. Note that the IS scheme in this corpus is different 
from the one in ISNotes in terms of fine-grained IS classes. In general, bridging in this corpus includes non-anaphoric, syntactically linked part-of and set-member relations (e.g., \emph{the house's door}), as well as comparative anaphors that are marked by surface indicators such as ``other'' or ``different''.\footnote{We refer the readers to \newcite{nissim04} for more details of fine-grained IS categories in the Switchboard dialogue IS corpus.} Nevertheless, we think the three main linguistic factors affecting a mention's IS analyzed in Section \ref{subsec:motivation} still hold in the dialogue domain.

\begin{table*}[t]
\begin{center}
\begin{tabular}{|p{2.8cm}|p{1.2cm}p{1.2cm}p{1.2cm}|p{1.2cm}p{1.2cm}p{1.2cm}|}
\hline
 &  \multicolumn{3}{c|}{\emph{SVM + Stanford Coreference}} &\multicolumn{3}{c|}{\emph{self-attention with}}  \\ 
 &  \multicolumn{3}{c|}{Rahman and Ng (2012)} &\multicolumn{3}{c|}{\emph{BERT$_{LARGE}$}} \\ 
 & R & P & F & R & P & F\\ \hline
\hline
 {old/ident} & 75.8&64.2&69.5&81.9&82.5&82.2\\
 {old/event}&2.4&31.8&4.5&73.2&71.5&72.3\\
 {old/general}&87.8&92.7&90.2&96.1&97.1&96.6\\
 {old/generic}&39.9&85.9&54.5&77.4&73.7&75.5\\
 {old/ident\_generic}&47.2&44.8&46.0&74.8&72.6&73.7\\
 {old/relative}&99.0&37.5&54.4&97.4&95.4&96.4\\
 {med/general}&84.0&72.2&77.7&85.3&74.6&79.6\\
 {med/bound}&2.7&40.0&5.1&29.7&43.4&35.3\\
 {med/part}&73.2&96.8&83.3&51.8&54.7&53.2\\
 {med/situation}&68.0&97.7&80.2&22.4&38.7&28.4\\
 {med/event}&46.3&100.0&63.3&16.2&22.9&19.0\\
 {med/set}&88.4&86.0&87.2&78.1&71.9&74.9\\
  {med/poss}&90.5&97.6&93.9&82.5&77.8&80.1\\
  {med/func\_value}&88.1&85.9&87.0&40.0&10.5&16.7\\
 {med/aggregation}&83.8&93.9&88.6&56.7&46.0&50.7\\
 {new}&90.4&83.6&86.9&67.9&76.9&72.1\\
\hline 
 acc &\multicolumn{3}{c|}{78.7} & \multicolumn{3}{c|}{\textbf{79.3}} \\
\hline
\end{tabular}
\end{center}
\caption{Results of our discourse context-aware self-attention model compared to the \emph{SVM + Stanford Coreference} model \cite{rahman12} on the Switchboard dialogue IS corpus for fine-grained IS classification.}
\label{tab:sw}
\end{table*}

Following \newcite{rahman12}, we split the dataset into a training set containing 117 dialogues and a testing set containing 30 dialogues. We train our model \emph{self-attention with BERT$_{LARGE}$} on the training dataset using the parameters described in Section \ref{subsec:parameter}. Table \ref{tab:sw} lists the results of our model compared to  \newcite{rahman12}'s system, which is an SVM$^{multiclass}$ model based on predictions from a rule-based system and the Stanford Deterministic Coreference  Resolution System \cite{leeheeyoung11}. The rule-based system consists of 18 hand-crafted rules for assigning IS subtypes to mentions. Some rules are based on the lexicon relations encoded in WordNet and FrameNet.

\begin{table*}[t]
\begin{center}
\begin{tabular}{|p{2.1cm}|p{12.9cm}|}
\hline
 \textbf{IS class}& \textbf{Most attended tokens} \\ \hline
 {old} & the, pre\_overlap2 = NA, pre\_overlap1 = NA, pre\_overlap2 = yes,
 pre\_overlap1 = yes, it, her, she, that, they\\ \hline
 {m/worldKnow.} & pre\_overlap1 = no, pre\_overlap2 = no, the, month, year, and, of,  to, this, said\\ \hline
 {m/syntactic}&pre\_overlap1 = no, the,  pre\_overlap2 = no, of, 's, her, in, its,  pre\_overlap2 = yes, to\\ \hline
 {m/aggregate} &and, the, or, pre\_overlap1 = no, pre\_overlap2 = yes, her, oil, units,  of, -\\ \hline
 {m/function}& \%, units, pre\_overlap2 = yes, pre\_overlap1 = no, to, 8, 
 fell,  5, 243,  million\\ \hline
 {m/comparative} &pre\_overlap1 = no, more, pre\_overlap2 = no, other, pre\_overlap2 = yes, higher, companies, some, of, that\\ 
\hline
 {m/bridging} &pre\_overlap1 = no, the,  pre\_overlap2 = no, a, in, friends, year, demand,  production, to\\ \hline
{new}&pre\_overlap1 = no, pre\_overlap2 = no, the, a, an, of, to, -, magazines, but \\

 \hline
\end{tabular}
\end{center}
\caption{Top ten most attended tokens for each IS class in \emph{self-attention with BERT$_{LARGE}$} trained on ISNotes.}
\label{tab:attentionmap}
\end{table*}

Note that the results of the two systems in Table \ref{tab:sw} are not directly comparable due to the different splits of the training/testing datasets.\footnote{ \newcite{rahman12} reported that their testing dataset contains 30 dialogues, but the split of the training/testing  datasets is not publicly available.} Neverthless, our model \emph{self-attention with BERT$_{LARGE}$} achieves competitive performance compared to \newcite{rahman12}'s system in terms of the overall accuracy. In general, it seems that our model is better at predicting \emph{old} mentions. We also checked the confusion matrix and found that the low results for \emph{med/situation}, \emph{med/event} and 
\emph{med/func\_value} is due to the fact that our model cannot distinguish these three categories from \emph{med/set}.

\section{Attention to Linguistic Features}
\label{sec:attention}
In order to gain additional insights into our model's predictions, we analyze the attention maps in our best model (\emph{self-attention with BERT$_{LARGE}$}) that is trained on ISNotes.
We aim to check to what extent the most attended tokens correspond to the linguistic features for each IS class. 

Specifically, we randomly choose one fold and apply the trained model to the testing dataset. Since the ``[CLS]'' token is used for prediction, we analyze the attention weights assigned to other tokens from ``[CLS]''  for each testing instance. The weight of each token is normalized by the sequence length. 
The final attended score for each token is calculated by aggregating normalized attended weights across all testing instances in all 16 heads from the last layer. This is because previous work suggests that the last layer usually encodes the task-specific features 
in fine-tuning \cite{kovaleva-etal-2019-revealing}.

Table \ref{tab:attentionmap} lists the top ten most attended tokens for each IS class. We exclude the separator tokens ([CLS]/[SEP]) and two punctuation tokens (comma and period) from the list, as 
suggested by \newcite{clark-etal-2019-bert} that these tokens are heavily attended in deep heads and might be used as a no-op for attention heads. Note that \emph{pre\_overlap1} and \emph{pre\_overlap2} are the two tokens that indicate whether the target mention has the same string/head with a mention from the preceding sentences. Both can have a value of ``yes'', ``no'', or ``NA''. Following \newcite{markert12}, ``NA'' means ``non-applicable'' and is mainly used for pronouns.

We notice that a lot of the attended tokens in Table \ref{tab:attentionmap} correspond well with the linguistic features for each IS class. For \emph{old} mentions, the model attends to pronouns and signals that indicate string overlap, while for \emph{new} mentions, the model attends to tokens that indicate string non-overlap and the indefinite determiners ``a/an''. 

It is interesting to note that the model seems to learn the internal syntactic/semantic structure for a few IS classes. For instance, ``of'' and ``'s'' are strong signals for \emph{m/syntactic} mentions that have a prepositional structure or a possessive structure. Also \emph{m/aggregate} mentions usually contain the tokens ``and/or'' that indicate the coordination structure. Similarly, for \emph{m/comparative} category,  the model learns to focus on a few premodifiers (e.g., ``more'', ``other'', and ``higher'') that indicate the comparison between two entities.

Finally for \emph{m/function} mentions, the model learns to mostly focus on numbers. Surprisingly, the model also learns to attend the verb ``\text{fell}'', which corresponds well with the definition of this IS class (see Section \ref{subsec:motivation}). For the most difficult category \emph{m/bridging}, it seems that the model attends to some relational nouns (e.g., ``friends'' or ``demand'') that are likely used as bridging anaphors.

\section{Conclusions}
We propose a simple discourse context-aware self-attention model for IS classification based on the BERT fine-tuning framework. 
We cast the IS classification problem as a sentence classification task by creating 
a novel ``pseudo sentence'' for each mention. We design the ``pseudo sentence''  based on the linguistic intuitions about IS and it contains most indicative context information to predict a mention's information status. 
Such design allows the model to capture both clues from the mention and its context when predicting IS. 

Our model does not contain any complex hand-crafted semantic features and achieves the new state-of-the-art results for IS classification and bridging anaphora recognition on ISNotes that contains written news articles. In another domain that consists of conversational dialogues (Switchboard), our model also achieves competitive performance for fine-grained IS classification compared to previous work \cite{rahman12}.

Finally, in order to better understand our model's predictions, we probe our best model (\emph{self-attention with BERT$_{LARGE}$}) on ISNotes. We find that our model learns to pay more attention to signals that correspond well to the linguistic features of each IS class.

\section*{Acknowledgments}
The author would like to thank Charles Jochim for proofreading the paper and the anonymous reviewers for their valuable feedback.


\bibliographystyle{coling}
\bibliography{../../bib/lit/lit}

\end{document}